\ifdefined\pdfminorversion\pdfminorversion=7\fi
\documentclass[11pt]{article}

\usepackage[T1]{fontenc}
\usepackage[utf8]{inputenc}
\usepackage{newtxtext,newtxmath}
\usepackage[margin=1in]{geometry}
\usepackage{microtype}
\usepackage{amsmath}
\usepackage{graphicx}
\usepackage{booktabs}
\usepackage{xcolor}
\usepackage[numbers,sort&compress]{natbib}
\usepackage[font=small,labelfont=bf]{caption}
\usepackage{enumitem}
\usepackage[hyphens]{xurl}
\usepackage[colorlinks=true,linkcolor=blue!50!black,citecolor=blue!50!black,urlcolor=blue!50!black]{hyperref}

\setlist{itemsep=2pt,topsep=4pt}
\title{\textbf{Site Is Decodable Before Pretraining:\\
Negative Controls for Probing Frozen Brain-MRI Foundation Models}}

\author{Saman Rahbar\\[2pt]
\normalsize Independent researcher, Vancouver, Canada\\
\normalsize \href{https://srahbar.com}{srahbar.com} \quad
\href{mailto:info@srahbar.com}{info@srahbar.com}}

\date{}

\begin{document}

\maketitle

\begin{abstract}
\noindent
Brain foundation models are often tested with a probe. The model is frozen, a
simple classifier is trained on its output, and the classifier's accuracy is
taken as a sign of what pretraining learned. We show that this reading can be
wrong unless two controls are reported with it. We probed three frozen 3-D
brain-MRI models at five depths, on two independent multi-site cohorts (ABIDE-I
and ABIDE-II). The site where a scan was acquired could be predicted at about
0.9, on a scale where 0 is chance and 1 is perfect. No clinical or demographic
target came close. An untrained copy of the same model
predicted site just as well on both cohorts. Untrained models of three
architectures, each built with three random seeds, did the same on ABIDE-I. The
raw image, shrunk to $12\times12\times12$ voxels and given to no model at all,
predicted site at 0.95. Pretraining did not create the site signal. It was already in the
scans. We also looked for any clinical target on which pretraining beats an
untrained model, and found no reliable case. Adjusting for age, sex and
diagnosis left site prediction almost unchanged, and so did a 4.6-fold increase
in the number of voxels. Resting-state fMRI from the same people showed the same
pattern. Site was the most predictable attribute of brain connectivity (0.81),
and it survived five untrained layers while age did not. We recommend reporting two
controls with every probe of a brain foundation model: running the probe again
on the untrained model, and on the raw input. We release the
code for both.
\end{abstract}

\section{Introduction}

Brain foundation models are trained on large collections of scans and then
reused without further training. The usual way to test such a model is a probe.
The model is frozen, and a simple classifier is trained to predict some
attribute from its output \citep{alain2016understanding}. If the classifier
predicts a clinical attribute well, this is read as evidence that pretraining
captured something useful.

A probe shows how well an attribute can be predicted. It does not show why.
There are three possible reasons. Pretraining may have learned the attribute,
the architecture may keep it even without training, or it may already be plain
in the input. Two simple controls tell these apart. The first runs the same probe
on the model with its weights left random, and the second runs it on the raw
input. Probing studies of language
models adopted controls like these years ago \citep{hewitt2019control}, and it
has long been known that untrained networks can produce useful features
\citep{saxe2011random}.

Brain foundation-model papers often leave out one or both controls. Two widely
used fMRI models are typical. BrainLM \citep{caro2024brainlm} compares its
clinical predictions with standard classifiers trained on connectivity, which is
a raw-input comparison. It does not test an untrained BrainLM. Brain-JEPA
\citep{dong2024brainjepa} compares its probes with those of BrainLM, also without
an untrained copy of itself. Neither paper reports how well the scanning site
can be predicted from its model's output. Without these numbers, a reader cannot
tell how much of a probe result is due to pretraining.

Acquisition site is a natural test case. Sites differ in scanner make, magnetic
field strength, scan settings and image reconstruction. These differences are
visible in the images and in features from trained networks
\citep{glocker2019machine,wachinger2021detect}. A large body of work, known as
harmonisation, removes site effects from measurements taken from the images
\citep{johnson2007combat,fortin2018harmonization}. Foundation models are now
used as general summaries of brain anatomy
\citep{tang2022self,cox2024brainsegfounder}, but their outputs are seldom
checked for site.

We ran that check on frozen brain-MRI models, and then ran the two controls.
Site could be predicted at about 0.9, far above any clinical target. An
untrained model predicted site just as well, and the raw image did about as
well. Without the controls, our first result would have looked like a finding
about what pretraining learns. With them, it is a finding about the scans.

\paragraph{Contributions.}
\begin{itemize}
\item We measure how well site and each clinical or demographic attribute can be
      predicted from three frozen brain-MRI models, at five depths, on two
      independent cohorts (Section~\ref{sec:matrix}).
\item Two controls show that the site signal does not come from pretraining. An
      untrained model matches the pretrained ones, across three architectures
      and three seeds. The raw image predicts site about as well as any model
      (Sections~\ref{sec:random} and~\ref{sec:raw}).
\item We search every target and depth for an advantage from pretraining and
      find none that holds up over repeated data splits. We also show that a
      single split can suggest an advantage that is not there
      (Section~\ref{sec:positive}).
\item We rule out three other explanations: differences between the people each
      site scanned, image resolution, and the type of classifier
      (Sections~\ref{sec:population} to~\ref{sec:resolution}).
\item We repeat the study on resting-state fMRI from the same people
      (Section~\ref{sec:fmri}).
\item We measure what removing the site signal costs. It is harmless for a
      single-label classifier and damaging for segmentation
      (Section~\ref{sec:removal}).
\item We recommend two controls for every probing study of brain foundation
      models, and release code for both (Section~\ref{sec:discussion}).
\end{itemize}

\begin{figure}[t]
\centering
\includegraphics[width=\textwidth]{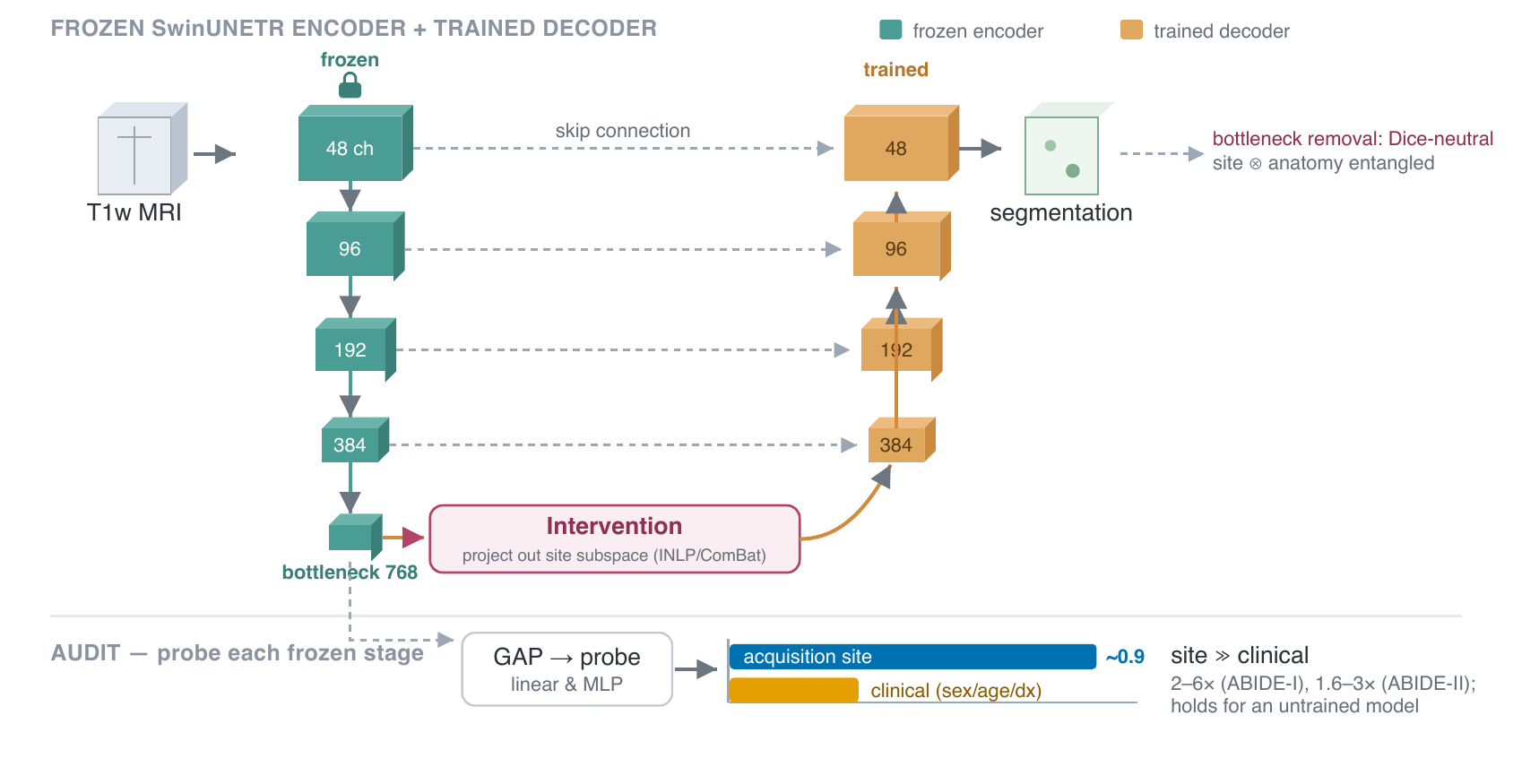}
\caption{\textbf{Study design.} The frozen SwinUNETR model (teal) has five
stages, from 48 to 768 channels. A decoder (amber) is trained only for the
segmentation experiment. \emph{Audit:} the output of each frozen stage is
averaged over space and probed for site and for each clinical or demographic
attribute. \emph{Controls:} the same probes on an untrained model and on the raw
image. \emph{Removal:} the site directions are removed from the frozen output
before decoding. ``Site'' covers the scanner, the scan settings and the people
each centre recruited.}
\label{fig:pipeline}
\end{figure}

\section{Related work}

\paragraph{Site and scanner effects.}
Brain scans pooled from several sites carry differences that come from the
equipment rather than the brain. These differences affect measurements such as
cortical thickness \citep{fortin2018harmonization,pomponio2020harmonization}.
Site can also be predicted from the features of trained deep networks
\citep{glocker2019machine}. The closest earlier study is
\citet{wachinger2021detect}. They pooled 35{,}320 brain scans from 17 studies
and could tell which study a scan came from with 71.5\% accuracy. They also
noted that harmonisation can remove information about the individual, which our
segmentation experiment measures directly. Sites also differ in whom they scan.
\citet{yamashita2019harmonization} separated the two sources by scanning the
same people at several sites. They found that both the equipment differences and
the differences in recruited people were comparable in size to the psychiatric
disorder effects under study. ABIDE has no such repeat scans, so we can only
adjust for the attributes that were recorded.

\paragraph{Harmonisation.}
ComBat \citep{johnson2007combat}, adapted for brain imaging by
\citet{fortin2018harmonization}, adjusts the mean and spread of each measurement
for each site. Later versions handle nonlinear age effects
\citep{pomponio2020harmonization} and site differences in how measurements vary
together \citep{chen2022covbat}. The second of these shows that adjusting means
and spreads alone can leave site predictable by a linear classifier. Other
methods remove scanner information while a network is trained, using an
adversarial objective \citep{dinsdale2021unlearning} or an information-based
objective \citep{moyer2020scanner}. All of these assume either hand-made
measurements or the chance to train the model. A frozen foundation model offers
neither. We therefore test a removal step applied after the model.

\paragraph{Shortcuts.}
Models often rely on features that predict the label without being the signal
of interest \citep{geirhos2020shortcut}. In medical imaging,
\citet{gichoya2022race} showed that networks can recognise a patient's
self-reported race from X-rays and CT scans. Measured confounders did not
explain this, and it survived heavy degradation of the images. We apply the same
three checks to site. We show that site can be predicted, that measured
attributes of the people do not explain it, and that it is present in a degraded
version of the image. \citet{seyyedkalantari2021underdiagnosis} show why such
hidden signals matter for patients.

\paragraph{Probing and its controls.}
Linear probes are the standard way to ask what a model's output contains
\citep{alain2016understanding}. \citet{hewitt2019control} showed that probe
accuracy is hard to interpret on its own. A flexible enough probe can learn
labels that the model does not represent in any useful way. Untrained networks
also give useful features, because the architecture itself shapes the output
\citep{saxe2011random}. We use both ideas. An untrained model is the control for
pretraining, and the raw input is the control for the model. To remove site we
use iterative null-space projection \citep{ravfogel2020null}. It repeatedly
trains a linear site classifier and removes the directions that classifier uses.

\paragraph{Brain foundation models.}
Self-supervised 3-D models trained on large collections of MRI and CT scans are
reused as frozen feature extractors
\citep{bommasani2021opportunities,tang2022self,cox2024brainsegfounder}. Many are
built on the SwinUNETR architecture \citep{hatamizadeh2022swinunetr}. Reviews of
medical foundation models list robustness and evaluation as open problems
\citep{moor2023foundation,zhang2024challenges}. For fMRI, BrainLM
\citep{caro2024brainlm} and Brain-JEPA \citep{dong2024brainjepa} test frozen and
fine-tuned models on clinical and demographic targets, as described above. We
apply both controls to frozen brain-MRI models, and add checks across depth,
architecture and random seed.

\section{Methods}
\label{sec:methods}

\subsection{Data and preprocessing}
We use T1-weighted MRI from ABIDE-I (546 people, 6 sites) and ABIDE-II
(989 people, 15 sites) \citep{dimartino2014abide,dimartino2017abideii}. The six
ABIDE-I sites (CALTECH, NYU, PITT, UM\_1, USM, YALE) and the six matched
ABIDE-II sites (BNI\_1, EMC\_1, ETHZ\_1, GU\_1, IP\_1, IU\_1) have none in
common. Every ABIDE-II result is therefore a repeat at new centres. Scans come from the public FCP-INDI mirror. Site, sex, age and autism
diagnosis come from the cohort tables.

We keep preprocessing to a minimum, matching what a user would feed a frozen
model. Each scan is reoriented to a standard direction (RAS), resampled to
1.7\,mm voxels, scaled to the range 0 to 1, and cropped or padded to $96^3$
voxels. We do not remove the skull, correct intensity bias, or align scans to a
template. For the resolution check we use 1.0\,mm voxels and $160^3$ volumes and
change nothing else. We exclude a person if age is missing, sex is not recorded,
the scan fails to load, or the model output contains invalid values.

\emph{Site} in this paper means the acquisition-site label. It combines the
scanner, the scan settings and the people each site recruited.
Section~\ref{sec:population} tests how much of the site signal comes from the
people.

\subsection{Models and their outputs}
We test three SwinUNETR models \citep{hatamizadeh2022swinunetr}. They have the
same architecture and differ only in their weights:
\begin{itemize}
\item \emph{brain-pretrained}: trained without labels on about 41{,}000
      UK Biobank scans \citep{cox2024brainsegfounder};
\item \emph{CT-pretrained}: the public self-supervised SwinUNETR from MONAI
      \citep{tang2022self};
\item \emph{untrained}: the same architecture with random weights from a fixed
      seed.
\end{itemize}
We use only the encoder, the part of the network that turns a scan into
features. No weight is changed anywhere in this paper. The encoder has five
stages, $L0$ to $L4$, with 48, 96, 192, 384 and 768 channels. At each stage we
average the output over space, which gives one vector per person.

The brain-pretrained model was trained on UK Biobank and is tested here on ABIDE.
That is the intended use. Foundation models are meant for data they have not
seen.
The untrained control uses no training data at all, so this difference in data
cannot affect it.

\subsection{Probes and the score}
Each output dimension is standardised before probing. Site, sex and autism
diagnosis are categories. For these we train a logistic regression classifier
($C=1$) on 70\% of the people and test it on the other 30\%, keeping category
proportions equal in both parts. We report balanced accuracy corrected for
chance:
\begin{equation}
d = \max\!\left(0,\; \frac{\mathrm{bAcc} - 1/K}{1 - 1/K}\right),
\label{eq:decodability}
\end{equation}
where $K$ is the number of categories. A score of 0 means guessing and 1 means
perfect. The correction makes a six-category target such as site roughly
comparable with a two-category target such as sex. Age is a number, so we fit a
ridge regression ($\alpha=1$) and report $R^2$, the share of age variation the
probe explains, also floored at 0. $R^2$ and the corrected accuracy both run from
0 to 1, but they measure different things. We therefore report every target
separately and never merge them into one score.

We repeat every probe with a small neural network (one hidden layer of 64 units,
$\alpha=10^{-3}$, at most 300 iterations) on the same data. If the network does
no better than the linear probe, the attribute is linearly predictable and the
linear score is not an underestimate. Table~\ref{tab:hyper} lists all settings.

\subsection{Statistics}
The main tables (Tables~\ref{tab:abide1}, \ref{tab:abide2} and
\ref{tab:resolution}) and Figure~\ref{fig:targets} report the mean over 50
random train and test splits. Intervals cover the middle 90\% of those 50
results. The control experiments use one fixed split instead: the raw-image
probe, the ViT and ResNet models, the seed sweep, the adjustment for the people
scanned, and the fMRI part. The main tables show that one split can move a site
score by about $\pm0.03$ to $\pm0.05$. Single-split values should be read with
that margin. We skip a cell if any category has fewer than four people.

To compare untrained and pretrained models we use 200 matched splits. A
one-sided sign test checks whether the untrained model is worse. A two
one-sided tests (TOST) procedure checks whether the two are equal within 0.05.
These $p$-values describe variation across data splits. They do not describe
variation across pretrained models, of which we have two.

\subsection{Controls}
\paragraph{Untrained models.} Beyond the untrained SwinUNETR, we build untrained
versions of a plain vision transformer (ViT) and a 3-D ResNet, and probe them the
same way. Each architecture is built with three seeds (0, 1 and 2). To keep this
affordable, the sweep uses at most 60 people per site, which leaves 331 people
from ABIDE-I. Its values therefore differ slightly from the main tables, so we
compare within each table only.
\paragraph{Raw image.} We probe site, sex and age directly from the image
shrunk to $12^3$ voxels (1{,}728 values), with no model. We use the same 331
people as the seed sweep.
\paragraph{Search for a pretraining advantage.} A control is most convincing when
the same comparison could also come out in favour of pretraining. We therefore
compare each pretrained model with the untrained one for sex, age and diagnosis
at every depth, which gives 30 comparisons. We count those where pretraining is
ahead by at least 0.05, the upper end of the split-to-split variation in our
tables. We run this on the 50-split results and, for comparison, on one split.
\paragraph{People scanned.} We remove the effects of age, sex and diagnosis from
the model output and measure site again.

\subsection{Resting-state fMRI}
For fMRI we use the preprocessed ABIDE region time series, with the CC200 brain
map \citep{craddock2012whole}, band-pass filtering and no global signal
regression. Matching to six sites leaves 534 people. The raw input is each
person's connectivity matrix: the correlation between every pair of regions,
Fisher-$z$ transformed and flattened into a vector. It has no learned parts. The
untrained model is a stack of five random linear layers, each 768 units wide,
with a ReLU after each layer. The width stays constant, so any drop with depth is
not caused by squeezing the data into fewer dimensions. We use three seeds and
one split per seed. We did not test a pretrained fMRI model.

\subsection{Removing the site signal}
We remove the linearly predictable site directions in two ways, both applied to
the frozen output. Iterative null-space projection (INLP)
\citep{ravfogel2020null} repeatedly trains a logistic site classifier and removes
the directions it relies on. We try 1, 2, 4, 8 and 12 rounds. ComBat
\citep{johnson2007combat,fortin2018harmonization} adjusts the mean and spread
for each site while keeping age and the target. As a control, we remove the same
number of random directions.

For single-label prediction, we train on all sites but one and test on the
remaining site, repeating this for every site. We test the change in each fold
with a sign test, a Wilcoxon test and a 300-sample bootstrap. For segmentation,
we attach a SwinUNETR decoder to the frozen encoder and train only the decoder,
using SynthSeg labels \citep{billot2023synthseg}. We then remove the site
directions from the feature maps as the scan passes through the network. We
report Dice, the overlap between predicted and reference labels (1 is perfect),
on held-out sites before and after.

\subsection{Implementation}
Only feature extraction uses a GPU. All probes run on a CPU with scikit-learn,
and the models use MONAI and PyTorch. The structural-MRI scripts include tests on
synthetic data that run without any real data. The loaders stop with an error if
data or model files are missing, rather than filling in defaults.

\section{Results}
\label{sec:results}

\subsection{Site is the most predictable attribute at every depth}
\label{sec:matrix}
\begin{figure}[t]
\centering
\includegraphics[width=\textwidth]{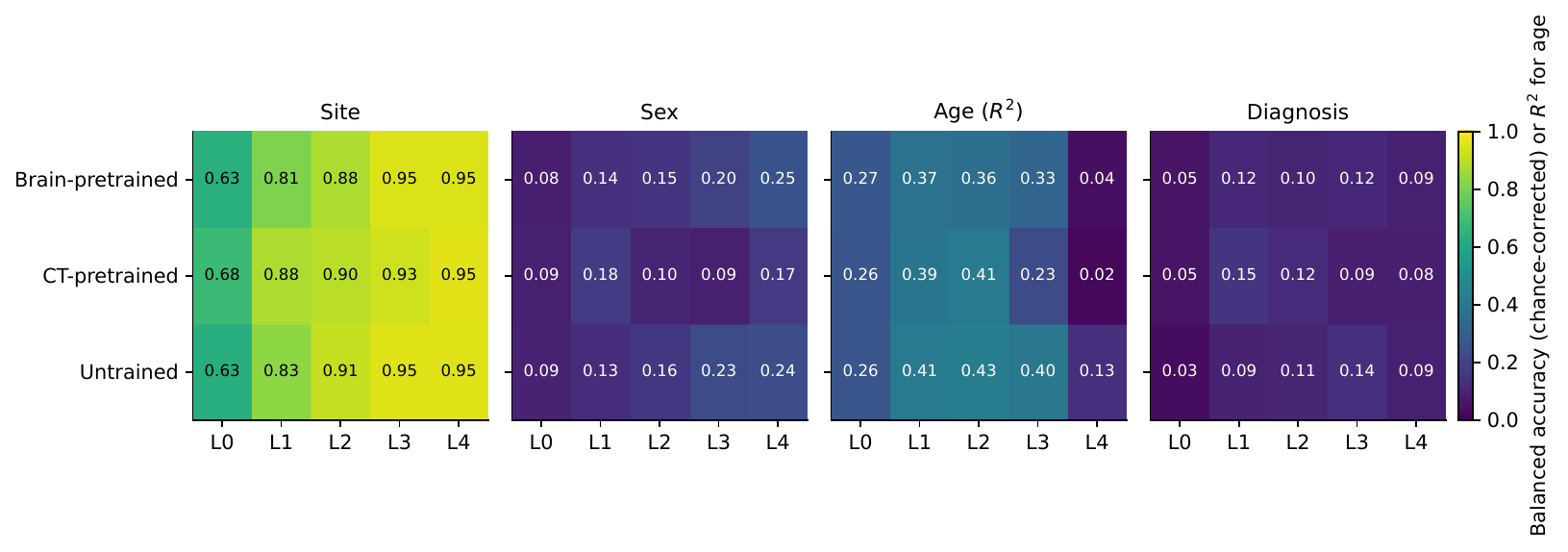}
\caption{\textbf{Site and each clinical or demographic target, by model and
depth} (ABIDE-I, 50 splits). Values come from the 1.0\,mm run of
Section~\ref{sec:resolution}. They agree closely with the 1.7\,mm run in the deep
layers and differ by up to 0.08 at $L0$ (Tables~\ref{tab:abide1}
and~\ref{tab:resolution}). Rows are models; columns are depths $L0$ to $L4$.
Site, sex and diagnosis are chance-corrected balanced accuracy
(Eq.~\ref{eq:decodability}); age is $R^2$. No clinical or demographic target
comes close to site at any depth.}
\label{fig:targets}
\end{figure}

On ABIDE-I, site scores range from 0.68 to 0.96 across the three models and five
depths (Figure~\ref{fig:targets}, Table~\ref{tab:abide1}). The 90\% intervals
are $\pm0.03$ to $\pm0.05$. The highest clinical or demographic score at any
depth is 0.41. Site scores rise with depth, from about 0.70 at $L0$ to about 0.95
at $L4$, for the brain-pretrained and untrained models, while the CT-pretrained model
levels off after $L3$. The small neural network is within 0.02 of the
linear probe everywhere, so site is linearly predictable and the linear scores
do not understate it.

ABIDE-II gives the same picture. In the matched six-site comparison, site scores
in the two deepest layers are 0.79 to 0.88. At $L4$ they are 0.87
$[0.81, 0.93]$ for the brain-pretrained model, 0.79 $[0.69, 0.87]$ for the
CT-pretrained model and 0.88 $[0.82, 0.93]$ for the untrained one. Clinical
scores are higher on this cohort, peaking near 0.5 in the middle layers, but
stay below site at every depth (Table~\ref{tab:abide2}). So the gap between site
and the clinical targets depends on the cohort, but on both cohorts site sits
near 0.9.

\subsection{An untrained model predicts site as well as the pretrained ones}
\label{sec:random}
\begin{figure}[t]
\centering
\begin{minipage}[t]{0.41\textwidth}
  \centering
  \includegraphics[width=\linewidth]{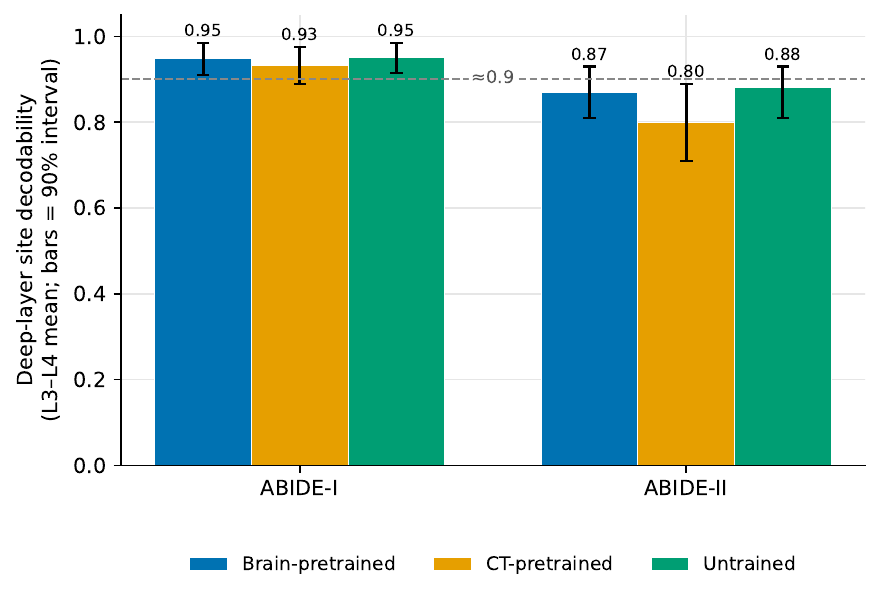}
  \caption{\textbf{Untrained versus pretrained.} Site scores averaged over the
  two deepest layers ($L3$ and $L4$). The untrained model matches the pretrained
  models on both cohorts. Error bars are 90\% intervals.}
  \label{fig:random}
\end{minipage}\hfill
\begin{minipage}[t]{0.55\textwidth}
  \centering
  \includegraphics[width=\linewidth]{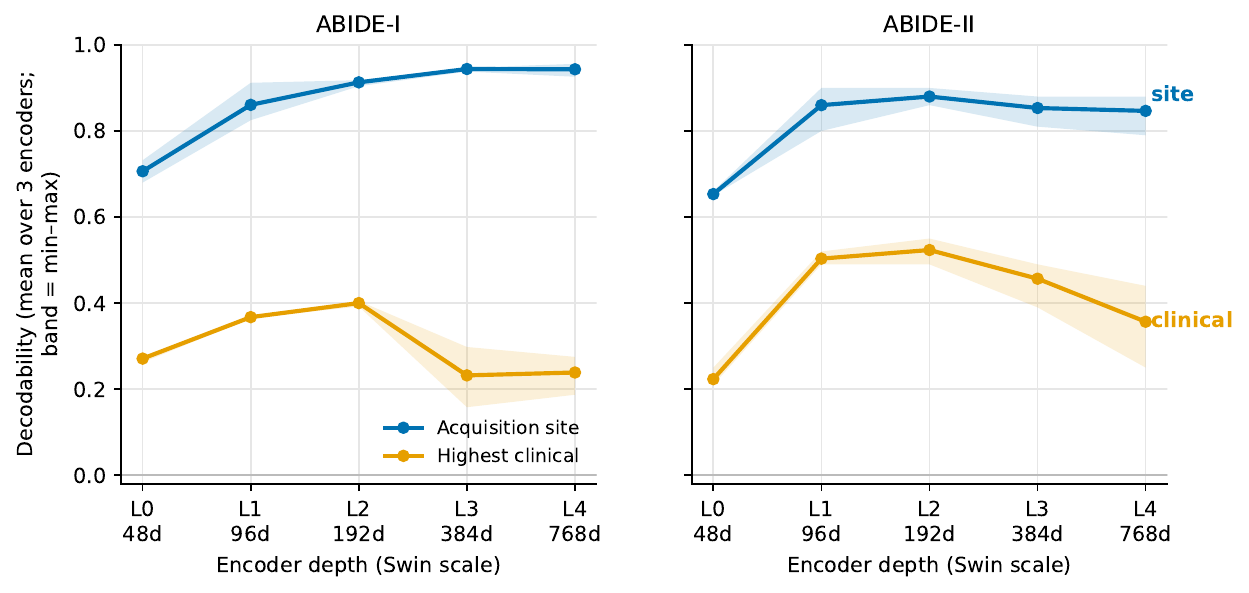}
  \caption{\textbf{Depth profiles} (1.7\,mm). Site against the highest of
  the sex, diagnosis and age scores at each depth, averaged over the three
  models; the band shows the range. The clinical curve mixes measures because
  it follows whichever target is highest. Figure~\ref{fig:targets} shows each
  target separately.}
  \label{fig:depth}
\end{minipage}
\end{figure}

On ABIDE-I the untrained SwinUNETR scores 0.956 $[0.92, 0.99]$ for site at $L4$,
against 0.948 for the brain-pretrained model and 0.926 for the CT-pretrained one
(Figure~\ref{fig:random}). On ABIDE-II the order is the same: 0.88, 0.87 and
0.79. Across 200 matched splits, the untrained model is no worse than either
pretrained model, with mean differences of $+0.007$ and $+0.018$ in its favour
(one-sided sign test, $p<10^{-17}$). The models are equal within 0.05 (TOST,
$p<10^{-50}$). These tests compare two specific pretrained models with their
untrained counterpart. For these two, pretraining is not needed for the site
signal and does not increase it. The tests do not show that pretraining never
adds site information.

The result does not depend on one architecture or one random draw. Untrained ViT
and 3-D ResNet models also predict site well in their deep layers
(Table~\ref{tab:arch}). Over three seeds per architecture, the $L4$ site score is
$0.973\pm0.006$ for SwinUNETR, $0.916\pm0.030$ for ViT and $0.987\pm0.019$ for
ResNet (Table~\ref{tab:seed}). Changing the seed moves the score no more than
changing the data split does.

\subsection{The raw image predicts site about as well as the models}
\label{sec:raw}
\begin{figure}[t]
\centering
\includegraphics[width=\textwidth]{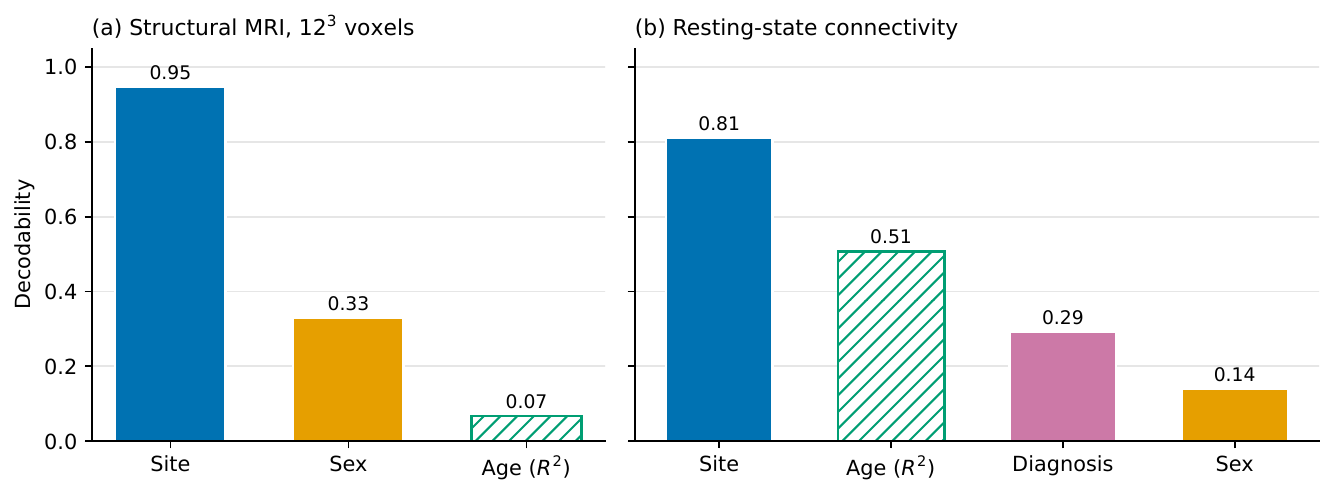}
\caption{\textbf{What the raw input already contains.} Scores for each target
from inputs that pass through no trained model. (a) Structural MRI shrunk to
$12^3$ voxels (ABIDE-I). (b) Resting-state connectivity matrices (ABIDE, six
matched sites, 534 people). Hatched bars are age $R^2$; solid bars are
chance-corrected balanced accuracy. One split each. Site is the most predictable
attribute of both inputs.}
\label{fig:raw}
\end{figure}

The image shrunk to $12^3$ voxels, with no model, predicts site at 0.95, while the
same input predicts sex at only 0.33 and age at $R^2=0.07$
(Figure~\ref{fig:raw}a).
For the same people, untrained models score 0.92 to 0.99 at $L4$
(Table~\ref{tab:seed}). The site signal is therefore present in coarse image
statistics before any network sees the scan. The models, pretrained or not, add
little to it. The scan itself carries the site.

\subsection{No reliable pretraining advantage on the clinical targets}
\label{sec:positive}
We searched sex, age and diagnosis at every depth for places where a pretrained
model beats the untrained one. We used the 50-split ABIDE-I results from the
1.0\,mm run, shown in Figure~\ref{fig:targets}. Pretraining is ahead by at least
0.05 in one of 30 comparisons: diagnosis at $L1$, where the CT-pretrained model
scores 0.153 $[0.06, 0.25]$ against 0.094 $[0.00, 0.20]$. The intervals overlap, and
the untrained model is ahead by at least 0.05 in nine comparisons. For these two
pretrained models, then, no measured target shows a reliable benefit from
pretraining.

A single split tells a different story. On one split of the same cohort, five of
30 comparisons favour pretraining by at least 0.05. The largest is diagnosis at
$L1$: 0.241 for the brain-pretrained model against 0.084 for the untrained one.
Over 50 splits, the same comparison differs by only 0.023. Clinical probes on a
cohort of this size are noisy enough that one split can show a pretraining
advantage that is not real. This is another reason to report repeated splits together with the
untrained control.

This result also limits what we can conclude. All models predict the clinical
targets weakly, so we cannot show that this comparison would detect a real
pretraining advantage. The site result does not rest on that. Every model
predicts site at about 0.9, and the raw image predicts it at 0.95. That
pretraining adds nothing to site is measured directly; it is not inferred from
how sensitive the comparison is.

\subsection{The people scanned do not explain the site signal}
\label{sec:population}
Each site recruited its own group of people, so the site signal could reflect
who was scanned rather than how. We removed the effects of age, sex and diagnosis
from the model output. The site score in the deep layers kept 95\% to 99.8\% of
its value. The brain-pretrained model stayed at 0.94, the CT-pretrained model went from 0.91
to 0.90, and the untrained one from 0.98 to 0.93.
This analysis uses one split, so its values differ slightly from
Table~\ref{tab:abide1}. The recorded attributes of the people explain very little
of the site signal. Unrecorded factors, such as head movement during the scan,
could still contribute (Section~\ref{sec:limitations}).

\subsection{Where site and clinical information sit in the network}
On ABIDE-I, age is best predicted in the middle of the network and poorly at the
last stage. In the 50-split values of Figure~\ref{fig:targets}, age $R^2$ is
0.36 to 0.43 at $L2$ and 0.02 to 0.13 at $L4$. Sex and diagnosis are weak at
every depth. On ABIDE-II, which carries more demographic signal, clinical scores
are higher and also peak at $L1$ to $L2$. Site behaves differently in the two
cohorts. It rises with depth on ABIDE-I but is flat or slightly falling after
$L2$ on ABIDE-II. Both cohorts share the same order: site is above every clinical
target at every depth, and clinical information, where present, is strongest in
the middle layers (Figure~\ref{fig:depth}). On ABIDE-I, the deepest layers are
the ones most often reused. They carry the most site information and the least
clinical information.

\subsection{Higher resolution does not change the result}
\label{sec:resolution}
At 1.7\,mm and $96^3$ voxels, the images might lack the detail needed to show
anatomy. That would lower the clinical scores and exaggerate the gap. We repeated
the whole study at 1.0\,mm and $160^3$ voxels, 4.6 times as many voxels, with
everything else unchanged (Table~\ref{tab:resolution}). Site scores in the deep
layers are 0.93 to 0.95, against 0.93 to 0.96 at 1.7\,mm. The best clinical score
rises only from 0.41 to 0.43, and site stays above every clinical target at every
depth for every model. The small neural network is within 0.03 of the linear probe in
all 15 cells. The untrained model still matches the pretrained ones, at 0.95 for
both $L3$ and $L4$.

\subsection{The same pattern in resting-state fMRI}
\label{sec:fmri}
A structural scan is a single image per person. To test a signal that changes
over time, we repeated the study on resting-state fMRI, using scans of the same
ABIDE cohort. Site is again the most predictable attribute of the raw connectivity
matrix, at 0.810. Age reaches $R^2=0.507$, diagnosis 0.293 and sex 0.139
(Figure~\ref{fig:raw}b). The site signal survives untrained layers. It scores
0.65 after one random layer and 0.36 after five, averaged over three seeds
(Table~\ref{tab:fmri}). Age does not survive: its $R^2$ is at most 0.004 at every
random layer and seed, apart from one value of 0.16 (first layer, one seed).
Random layers therefore discard the age information in connectivity but keep
much of the site information.

Two differences from the structural results stand out. First, site leads the
best clinical target by less: about $1.6\times$ instead of $2$ to $6\times$.
Connectivity carries much more age information than a single structural image.
Second, depth has the opposite effect. Site scores rise with depth in the
structural models (0.70 to 0.95) and fall with depth in the untrained
connectivity model (0.65 to 0.36). Both kinds of data show the problem, but its
size and its pattern across depth differ. The controls therefore need to be run
for each model, not borrowed from another study. We did not test a pretrained
fMRI model, so this part compares the raw input with untrained layers only. It
shows that site is in the input and survives an untrained model, but not that
pretraining adds nothing for fMRI.

\subsection{Removing the site signal: cheap for single-label tasks, costly for segmentation}
\label{sec:removal}
The site directions can be removed from the frozen output after the fact. INLP
lowers the site score at the last stage from 0.94 to 0.066, and ComBat lowers it
to 0.000. For single-label prediction on held-out sites, the better method
depends on the target. For sex, which is largely separate from site, INLP is slightly
ahead of ComBat in balanced accuracy (0.626 against 0.616, and 0.580 with no
removal). For autism diagnosis, which is tied to site, ComBat does better because
it keeps the target while adjusting. The gap between 0.626 and 0.616 is within
the variation between splits. Neither method gives a significant gain on held-out
sites, because the averaged output carries little clinical information to begin
with.

Segmentation behaves differently. Removing the site directions at the last stage
alone leaves Dice unchanged (mean change about $2\times10^{-5}$). Removing them
at all five stages drops Dice on held-out sites from 0.84 to 0.18. Removing the
same number of random directions leaves Dice at 0.83. The damage therefore comes
from the site directions themselves, not from losing that many dimensions: the
directions that carry site also carry anatomy. The loss is largest in deep brain
structures and the brainstem (Table~\ref{tab:region}, Figure~\ref{fig:region}).
Before removal, a decoder trained on only two sites reaches 0.84 on held-out
sites. A decoder with skip connections therefore copes with the site signal.
These Dice values are measured against SynthSeg labels, which are built to be
robust to scanner differences. They partly reflect that labelling tool and are
not an absolute accuracy.

\section{Discussion}
\label{sec:discussion}

\paragraph{What the controls change.}
Without controls, our first result reads as ``frozen brain-MRI foundation models
encode the scanning site strongly.'' With them, the accurate statement is
different. The site is strongly present in the scans, and the models, trained or
not, keep it. The two readings call for different fixes. If pretraining had
added the site signal, cleaner training data or a site-blind training objective
would remove it. In fact the signal comes from basic image properties that every
model we tested keeps. Site has to be removed on purpose, either after the frozen
model or by changing what goes into it. Training objectives that discard site
information \citep{dinsdale2021unlearning,moyer2020scanner} and training on
randomly varied scan settings, as SynthSeg does \citep{billot2023synthseg}, are
the main options.

\paragraph{A reporting recommendation.}
We recommend that probes of brain foundation models always be reported with two
extra numbers: the same probe on the untrained model, and on the raw input. Neither needs pretraining, extra
data or tuning. Here they cost a few GPU hours, and they changed the conclusion.
The untrained number shows whether pretraining helped. The raw-input number shows
whether the model was needed at all. A site probe in particular is a single
logistic regression and could be reported as routinely as a table of participant
ages and sexes.

\paragraph{Other practical points.}
On ABIDE-I the deepest layers carry the most site information and the least
clinical information. Deeper layers should not be assumed to hold more anatomy,
and the choice of layer should be tested. Shared or pooled model outputs reveal
their scanning site with scores of about 0.9 to 0.96. This reveals the site, not
the individual, and we did not test an attack on a real system. Removing site is
cheap and safe for single-label classifiers but can damage segmentation, so the
removal method should suit the downstream task.

\paragraph{Beyond MRI.}
Every attribute in this study has a counterpart in other body signals. In EEG
and MEG, site corresponds to the amplifier, electrode layout and session. In
wearables, it corresponds to where the sensor sits. These are properties of the
recording. They are present in the raw signal, and a model that represents the
signal faithfully will tend to keep them. When a probe for person, session or
device scores high, the score should not be credited to pretraining until both
controls have been run. The reverse also matters. A clinical prediction that
beats chance may be using a recording property that happens to be linked to the
label, which is the evaluation version of shortcut learning
\citep{geirhos2020shortcut}.

\paragraph{What the controls do not settle.}
Passing the untrained control does not make a model clinically useful, and
failing it does not make a model useless. The models we tested still work well
for the segmentation task they were built for. Our claim is about how results are
read: probe accuracy alone does not show what pretraining learned.

\subsection{Limitations}
\label{sec:limitations}
\begin{itemize}
\item \textbf{Pretrained models.} Both pretrained models are SwinUNETR. The
      untrained result holds across three architectures, but the comparison with
      pretraining covers only two pretrained models of one family. Testing more
      pretrained models is future work.
\item \textbf{People scanned.} Site combines the scanner, the settings and the
      people scanned. Adjusting for age, sex and diagnosis leaves the site score
      almost unchanged, but unrecorded factors such as head movement could
      contribute. Scanning the same people at several sites
      \citep{yamashita2019harmonization} would settle this. ABIDE does not
      include such scans.
\item \textbf{Sensitivity of the comparison.} We found no reliable pretraining
      advantage on any clinical target. We therefore cannot show that the
      untrained comparison would detect one here. The site result does not depend
      on this (Section~\ref{sec:positive}).
\item \textbf{Diagnosis on new sites.} Our clinical probes mix sites between
      training and testing; we hold out whole sites only in the removal
      experiment. Holding out sites for diagnosis would show whether pretraining
      helps on a site the model has never seen.
\item \textbf{Comparing targets.} The chance correction makes targets with
      different numbers of categories roughly comparable, but not exactly. The
      gap between site and the clinical targets is therefore approximate. Our two
      main results, untrained against pretrained and raw against model, compare
      site with site and do not depend on it.
\item \textbf{Cohorts.} Both cohorts come from autism studies and may not
      represent clinical brain imaging in general. The ABIDE-II results use a
      matched six-site subset.
\item \textbf{Probes.} A linear probe gives a lower bound on what can be
      predicted. The small neural network does no better here, but other
      nonlinear structure may exist. Probe accuracy can also reflect the probe's
      capacity as well as the model's content \citep{hewitt2019control}.
\item \textbf{fMRI.} The fMRI part has no pretrained model, so it compares the
      raw input with untrained layers only.
\item \textbf{Segmentation labels.} Dice is measured against SynthSeg labels,
      not manual tracings.
\end{itemize}

\section*{Code and data availability}
ABIDE-I and ABIDE-II are openly available from the ABIDE initiative
(\url{http://fcon_1000.projects.nitrc.org/indi/abide/}). The brain-pretrained
(BrainSegFounder) and CT-pretrained SwinUNETR weights are public, and the
untrained models use fixed seeds. All analysis code is available at
\url{https://github.com/saman-rahbar/scanner-fingerprints}. This includes the
main tables, both controls, the search for a pretraining advantage, the
adjustment for the people scanned, the fMRI part and the removal experiments.

\section*{Acknowledgements}
This work used computing resources provided by the Digital Research Alliance of
Canada. The author thanks Dr.\ Sidney Fels (University of British Columbia) for
access to those resources. No specific grant funding was received.

\section*{Ethics}
This study reuses public, de-identified MRI data collected under the original
studies' ethics approvals. No new data were collected.

\bibliographystyle{plainnat}
\bibliography{references}

\clearpage
\appendix
\section{Full tables}
\label{app:tables}

\begin{figure}[h]
\centering
\includegraphics[width=0.6\textwidth]{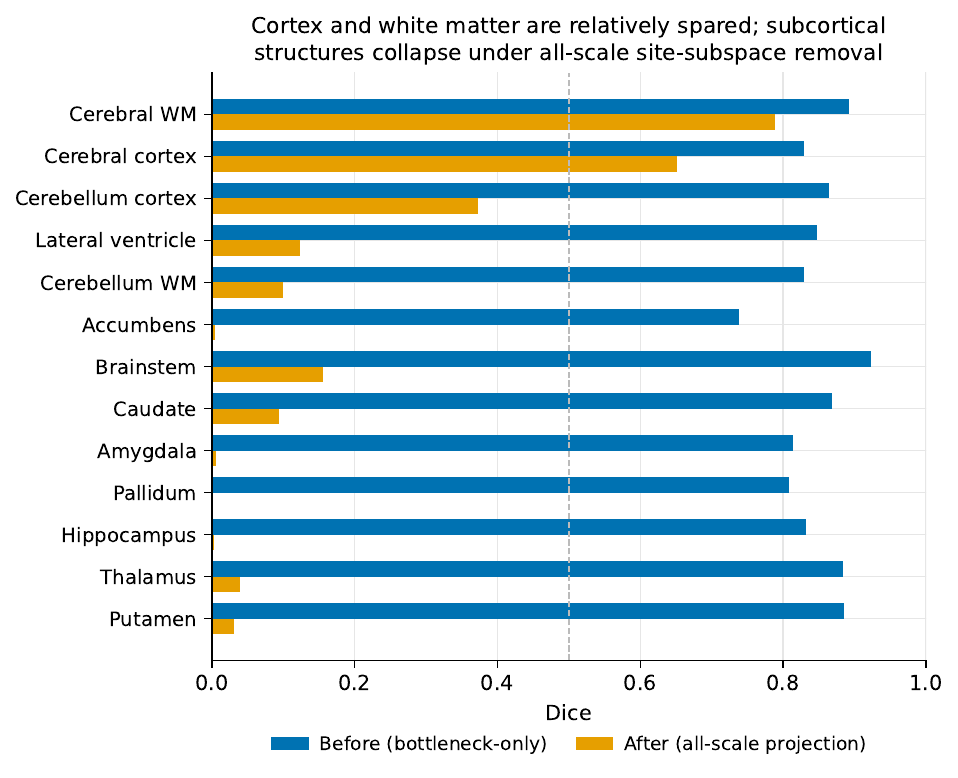}
\caption{\textbf{Per-region Dice before and after removing the site directions
at all scales}, sorted by change. Cortex and white matter lose little; every
subcortical structure and the brainstem fall close to zero.}
\label{fig:region}
\end{figure}

Tables~\ref{tab:abide1} and~\ref{tab:abide2} give site and the best clinical or
demographic value per model and depth; Figure~\ref{fig:targets} shows each
target separately. Tables~\ref{tab:arch} and~\ref{tab:seed} give the
architecture and seed controls, Table~\ref{tab:resolution} the resolution check,
Table~\ref{tab:fmri} the fMRI results, and Table~\ref{tab:region} with
Figure~\ref{fig:region} the per-region segmentation results.

\begin{table}[h]
\centering
\small
\caption{ABIDE-I site score (linear probe, chance-corrected) and the highest
clinical or demographic value at each depth. The clinical row takes the maximum
over sex, diagnosis (chance-corrected balanced accuracy) and age ($R^2$); see
Figure~\ref{fig:targets} for each target. The small neural-network probe is within 0.02 of
the linear site value in every cell.}
\label{tab:abide1}
\begin{tabular}{llccccc}
\toprule
Model & & $L0$ & $L1$ & $L2$ & $L3$ & $L4$ \\
\midrule
Brain-pretrained & site     & 0.71 & 0.83 & 0.90 & 0.95 & 0.95 \\
                 & highest clinical & 0.28 & 0.37 & 0.41 & 0.24 & 0.25 \\
CT-pretrained    & site     & 0.73 & 0.91 & 0.92 & 0.94 & 0.93 \\
                 & highest clinical & 0.26 & 0.37 & 0.39 & 0.16 & 0.19 \\
Untrained        & site     & 0.68 & 0.85 & 0.92 & 0.95 & 0.96 \\
                 & highest clinical & 0.28 & 0.37 & 0.40 & 0.30 & 0.28 \\
\bottomrule
\end{tabular}
\end{table}

\begin{table}[h]
\centering
\small
\caption{ABIDE-II, matched six-site comparison, means over repeated splits. The
nonlinear probe is within 0.03 of the linear site value in every cell.}
\label{tab:abide2}
\begin{tabular}{llccccc}
\toprule
Model & & $L0$ & $L1$ & $L2$ & $L3$ & $L4$ \\
\midrule
Brain-pretrained & site     & 0.65 & 0.80 & 0.88 & 0.87 & 0.87 \\
                 & highest clinical & 0.21 & 0.49 & 0.55 & 0.49 & 0.38 \\
CT-pretrained    & site     & 0.65 & 0.90 & 0.86 & 0.81 & 0.79 \\
                 & highest clinical & 0.21 & 0.52 & 0.53 & 0.39 & 0.25 \\
Untrained        & site     & 0.66 & 0.88 & 0.90 & 0.88 & 0.88 \\
                 & highest clinical & 0.25 & 0.50 & 0.49 & 0.49 & 0.44 \\
\bottomrule
\end{tabular}
\end{table}

\begin{table}[h]
\centering
\footnotesize
\begin{minipage}[t]{0.45\textwidth}
\centering
\caption{Site scores of untrained models of three architectures
(ABIDE-I). The SwinUNETR row repeats the untrained row of
Table~\ref{tab:abide1} ($n=546$, 50 splits); ViT and ResNet use $n=331$ and one
split.}
\label{tab:arch}
\begin{tabular}{lccccc}
\toprule
 & $L0$ & $L1$ & $L2$ & $L3$ & $L4$ \\
\midrule
SwinUNETR & 0.68 & 0.85 & 0.92 & 0.95 & 0.96 \\
ViT       & 0.91 & 0.90 & 0.91 & 0.90 & 0.93 \\
ResNet    & 0.95 & 0.95 & 0.98 & 0.95 & 0.96 \\
\bottomrule
\end{tabular}
\end{minipage}\hfill
\begin{minipage}[t]{0.51\textwidth}
\centering
\caption{$L4$ site scores of untrained models over three seeds
(ABIDE-I, $n=331$, one split per seed).}
\label{tab:seed}
\begin{tabular}{lcccc}
\toprule
 & seed 0 & seed 1 & seed 2 & mean $\pm$ sd \\
\midrule
SwinUNETR & 0.977 & 0.965 & 0.977 & $0.973\pm0.006$ \\
ViT       & 0.932 & 0.942 & 0.875 & $0.916\pm0.030$ \\
ResNet    & 0.960 & 1.000 & 1.000 & $0.987\pm0.019$ \\
\bottomrule
\end{tabular}
\end{minipage}
\end{table}

\begin{table}[h]
\centering
\small
\caption{Resolution check (ABIDE-I, 1.0\,mm, $160^3$, $n=546$), same protocol as
Table~\ref{tab:abide1}. ``MLP'' is the small neural-network site probe; ``ratio'' is site
divided by the highest clinical value. Intervals span $\pm0.03$ to $\pm0.07$ and
are in the released results.}
\label{tab:resolution}
\begin{tabular}{llccccc}
\toprule
Model & & $L0$ & $L1$ & $L2$ & $L3$ & $L4$ \\
\midrule
Brain-pretrained & site     & 0.63 & 0.81 & 0.88 & 0.95 & 0.95 \\
                 & MLP      & 0.64 & 0.81 & 0.90 & 0.95 & 0.95 \\
                 & highest clinical & 0.27 & 0.37 & 0.36 & 0.33 & 0.25 \\
                 & ratio    & 2.3  & 2.2  & 2.4  & 2.9  & 3.8  \\
\midrule
CT-pretrained    & site     & 0.68 & 0.88 & 0.90 & 0.93 & 0.95 \\
                 & MLP      & 0.68 & 0.89 & 0.92 & 0.94 & 0.96 \\
                 & highest clinical & 0.26 & 0.39 & 0.41 & 0.23 & 0.17 \\
                 & ratio    & 2.6  & 2.3  & 2.2  & 4.0  & 5.6  \\
\midrule
Untrained        & site     & 0.63 & 0.83 & 0.91 & 0.95 & 0.95 \\
                 & MLP      & 0.63 & 0.86 & 0.92 & 0.95 & 0.95 \\
                 & highest clinical & 0.26 & 0.41 & 0.43 & 0.40 & 0.24 \\
                 & ratio    & 2.4  & 2.0  & 2.1  & 2.4  & 4.0  \\
\bottomrule
\end{tabular}
\end{table}

\begin{table}[h]
\centering
\small
\caption{Resting-state fMRI (ABIDE, CC200, six matched sites, $n=534$), one
split per seed, means over three seeds. The connectivity model is the
untrained model of Section~\ref{sec:methods}, applied to the connectivity
vector. The temporal model is a random five-stage 1-D convolutional stack
applied to the region time series and averaged over time, shown for comparison;
pooling over time discards the correlations between regions, and its site
scores are correspondingly lower. Site, sex and diagnosis are
chance-corrected balanced accuracy; age is $R^2$.}
\label{tab:fmri}
\begin{tabular}{llccccc}
\toprule
Input & Target & \multicolumn{5}{c}{Layer} \\
 & & 1 & 2 & 3 & 4 & 5 \\
\midrule
Raw connectivity & site / age / diagnosis / sex & \multicolumn{5}{c}{0.810 / 0.507 / 0.293 / 0.139} \\
\midrule
Untrained connectivity model & site      & 0.65 & 0.56 & 0.49 & 0.46 & 0.36 \\
                            & age ($R^2$) & 0.05 & 0.00 & 0.00 & 0.00 & 0.00 \\
                            & diagnosis & 0.23 & 0.19 & 0.12 & 0.14 & 0.12 \\
                            & sex       & 0.12 & 0.10 & 0.03 & 0.03 & 0.05 \\
\midrule
Untrained temporal model     & site      & 0.42 & 0.28 & 0.19 & 0.12 & 0.10 \\
                            & age ($R^2$) & 0.02 & 0.00 & 0.00 & 0.00 & 0.00 \\
                            & diagnosis & 0.10 & 0.15 & 0.08 & 0.05 & 0.01 \\
                            & sex       & 0.09 & 0.06 & 0.09 & 0.06 & 0.04 \\
\bottomrule
\end{tabular}
\end{table}

\begin{table}[h]
\centering
\footnotesize
\begin{minipage}[t]{0.40\textwidth}
\centering
\caption{Per-region Dice for site removal at all scales (mean over four held-out
sites; decoder trained on two sites).}
\label{tab:region}
\begin{tabular}{lccc}
\toprule
Region & Before & After & Change \\
\midrule
Cerebral WM        & 0.892 & 0.788 & $-0.103$ \\
Cerebral cortex    & 0.830 & 0.651 & $-0.179$ \\
Cerebellum cortex  & 0.865 & 0.373 & $-0.492$ \\
Lateral ventricle  & 0.847 & 0.124 & $-0.723$ \\
Cerebellum WM      & 0.829 & 0.099 & $-0.730$ \\
Accumbens          & 0.738 & 0.004 & $-0.734$ \\
Brainstem          & 0.923 & 0.155 & $-0.768$ \\
Caudate            & 0.868 & 0.094 & $-0.774$ \\
Amygdala           & 0.814 & 0.007 & $-0.807$ \\
Pallidum           & 0.808 & 0.001 & $-0.808$ \\
Hippocampus        & 0.832 & 0.004 & $-0.828$ \\
Thalamus           & 0.884 & 0.040 & $-0.844$ \\
Putamen            & 0.885 & 0.031 & $-0.854$ \\
\bottomrule
\end{tabular}
\end{minipage}\hfill
\begin{minipage}[t]{0.57\textwidth}
\centering
\caption{Implementation settings. Splits are stratified for categorical targets;
scores and $R^2$ are floored at 0.}
\label{tab:hyper}
\begin{tabular}{ll}
\toprule
Setting & Value \\
\midrule
Orientation, spacing & RAS, 1.7\,mm (bilinear) \\
Input size, intensity & $96^3$, scaled to $[0,1]$ \\
Resolution check & 1.0\,mm, $160^3$ \\
Encoder & SwinUNETR, feature size 48 \\
Stage widths & 48/96/192/384/768 \\
Pooling, precision & spatial mean, fp32, no gradients \\
Split, repeats & 70/30, 50 repeats \\
Interval & 5th--95th percentile (90\%) \\
Logistic probe & $C=1$ \\
Ridge probe (age) & $\alpha=1$ \\
Nonlinear probe & 64 units, $\alpha=10^{-3}$, $\le300$ iter. \\
Minimum class count & 4 \\
INLP & logistic $C=1$, 1/2/4/8/12 iterations \\
ComBat covariates & age and target \\
Cross-site protocol & leave one site out \\
Bootstrap, paired splits & 300, 200 \\
TOST margin & 0.05 \\
Seeds (controls) & 0, 1, 2 \\
Subjects per site (controls) & 60 ($n=331$) \\
Raw-input probe & $12^3$ voxels \\
\bottomrule
\end{tabular}
\end{minipage}
\end{table}

\end{document}